\pdfoutput=1

\documentclass[11pt]{article}

\usepackage{tabularx}
\newcolumntype{Y}{>{\arraybackslash}X}

\usepackage[]{EMNLP2023}

\usepackage{times}
\usepackage{latexsym}

\usepackage[T1]{fontenc}

\usepackage[utf8]{inputenc}

\usepackage{microtype}

\usepackage{inconsolata}
\usepackage{musicography}
\usepackage{graphicx}
\usepackage{xcolor}
\usepackage[normalem]{ulem}
\useunder{\uline}{\ul}{}
\usepackage{multirow}
\usepackage[ruled]{algorithm2e}
\usepackage{arydshln}

\title{\textit{AGent}: A Novel Pipeline for Automatically Creating \\ Unanswerable Questions}

\author{Son Quoc Tran$^{1,3}$, Gia-Huy Do$^{1}$, Phong Nguyen-Thuan Do$^{3}$,\\ \textbf{Matt Kretchmar} $^{1}$\textbf{,} \textbf{Xinya Du}$^{2}$\\
$^1$Denison University \\ 
$^2$University of Texas at Dallas\\
$^3$The UIT NLP Group, Ho Chi Minh City\\
\texttt{\{tran\_s2, do\_g1, kretchmar\}@denison.edu}\\
\texttt{phongdntvn@gmail.com, xinya.du@utdallas.edu}}
\begin{document}
\maketitle
\begin{abstract}
The development of large high-quality datasets and high-performing models have led to significant advancements in the domain of Extractive Question Answering (EQA). This progress has sparked considerable interest in exploring unanswerable questions within the EQA domain. Training EQA models with unanswerable questions helps them avoid extracting misleading or incorrect answers for queries that lack valid responses. However, manually annotating unanswerable questions is labor-intensive. To address this, we propose \textit{AGent}, a novel pipeline that automatically creates new unanswerable questions by re-matching a question with a context that lacks the necessary information for a correct answer. In this paper, we demonstrate the usefulness of this \textit{AGent} pipeline by creating two sets of unanswerable questions from answerable questions in SQuAD and HotpotQA. These created question sets exhibit low error rates. Additionally, models fine-tuned on these questions show comparable performance with those fine-tuned on the SQuAD 2.0 dataset on multiple EQA benchmarks.
\footnote{Our code is publicly available at \url{https://github.com/sonqt/agent-unanswerable}.}
\end{abstract}

\section{Introduction}
\begin{figure}[ht]
\centering
\includegraphics[width =\linewidth]{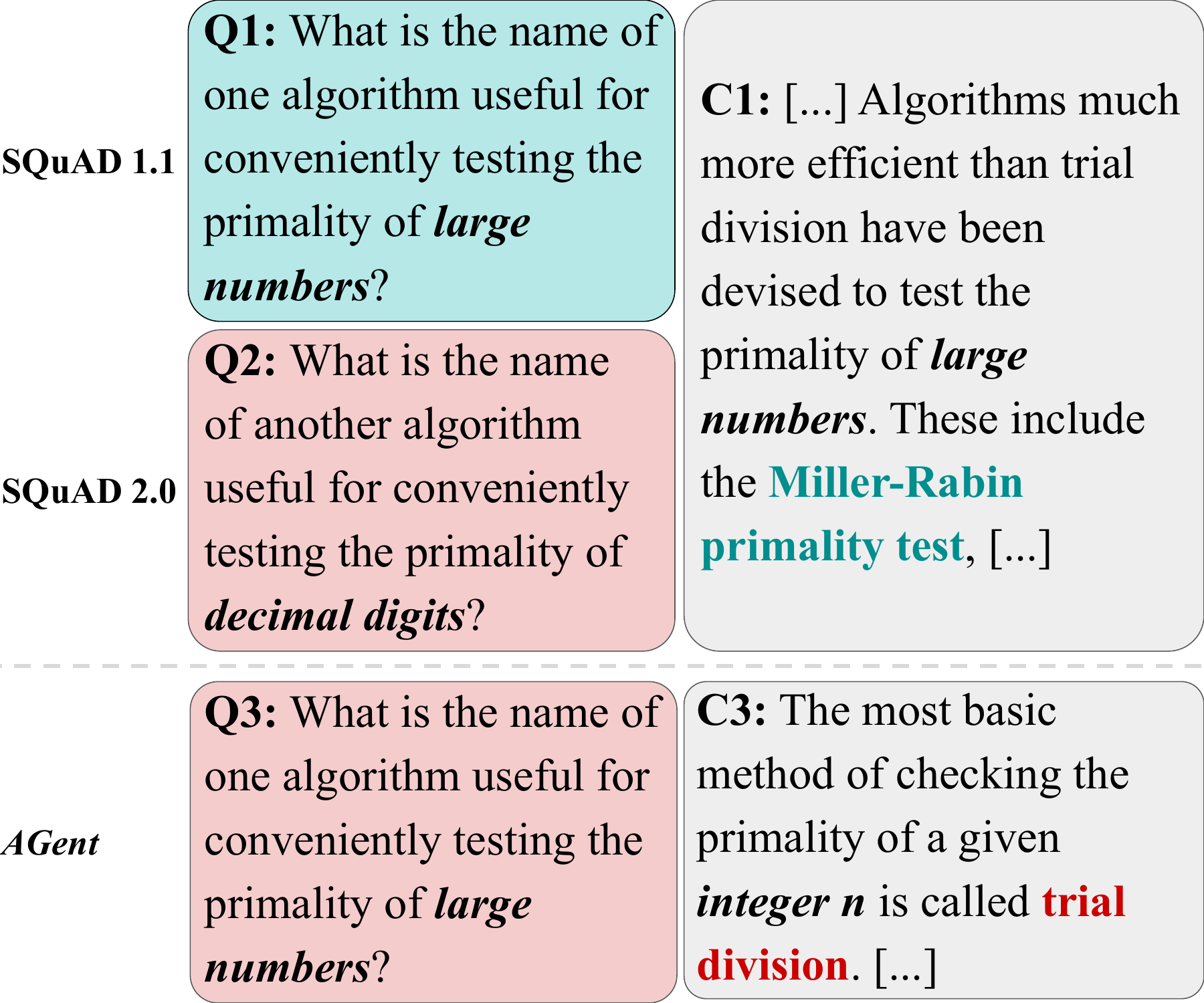}
\caption{Examples of an answerable question $Q1$ from SQuAD 1.1, and two unanswerable questions $Q2$ from SQuAD 2.0 and $Q3$ from SQuAD \textit{AGent}. In SQuAD 2.0,  crowdworkers create unanswerable questions by replacing ``large numbers'' with ``decimal digits.'' On the other hand, our automated \textit{AGent} pipeline matches the original question $Q1$, now $Q3$, with a new context $C3$. The pair $C3-Q3$ is unanswerable as context $C3$ does not indicate whether the \textcolor{red}{\textbf{trial division}} can \textbf{conveniently} test the primality of \textbf{large} numbers.}
\label{fig:example}
\end{figure}
Extractive Question Answering (EQA) is an important task of Machine Reading Comprehension (MRC), which has emerged as a prominent area of research in natural language understanding. Research in EQA has made significant gains thanks to the availability of many challenging, diverse, and large-scale datasets \cite{rajpurkar-etal-2016-squad, rajpurkar-etal-2018-know, kwiatkowski-etal-2019-natural, yang-etal-2018-hotpotqa, trivedi-etal-2022-musique}. Moreover, recent advancements in datasets also lead to the development of multiple systems in EQA \cite{huang2018fusionnet, NEURIPS2020_c8512d14} that have achieved remarkable performance, approaching or even surpassing human-level performance across various benchmark datasets.

Matching the rapid progress in EQA, the subfield of unanswerable questions has emerged as a new research area. Unanswerable questions are those that cannot be answered based only on the information provided in the corresponding context. Unanswerable questions are a critical resource in training EQA models because they allow the models to learn how to avoid extracting misleading answers when confronted with queries that lack valid responses. Incorporating unanswerable questions in the training set of EQA models enhances the overall reliability of these models for real-world applications \cite{tran-etal-2023-impacts}. 

Nevertheless, the manual annotation of unanswerable questions in EQA tasks can be prohibitively labor-intensive. Consequently, we present a novel pipeline to automate the creation of high-quality unanswerable questions given a dataset comprising answerable questions. This pipeline uses a retriever to re-match questions with paragraphs that lack the necessary information to answer them adequately. Additionally, it incorporates the concept of adversarial filtering for identifying challenging unanswerable questions. The key contributions of our work can be summarized as follows:
\begin{enumerate}
    \item We propose \textbf{\textit{AGent}} which is a novel pipeline for automatically creating unanswerable questions. 
    In order to prove the utility of \textit{AGent}, we apply our pipeline on two datasets with different characteristics, SQuAD and HotpotQA, to create two different sets of unanswerable questions. In our study, we show that the two unanswerable question sets created using \textit{AGent} pipeline exhibit a low error rate.
    \item Our experiments show that the two unanswerable question sets created using our proposed pipeline are challenging for models fine-tuned using human annotated unanswerable questions from SQuAD 2.0. Furthermore, our experiments show that models fine-tuned using our automatically created unanswerable questions show comparable performance to those fine-tuned on the SQuAD 2.0 dataset on various EQA benchmarks, such as SQuAD 1.1, HotpotQA, and Natural Questions. 
\end{enumerate}

\section{Related Work}
\subsection{Unanswerable Questions}
In the early research on unanswerable questions, \citet{levy-etal-2017-zero} re-defined the BiDAF model \cite{seo2018bidirectional} to allow it to output whether the given question is unanswerable. Their primary objective was to utilize MRC as indirect supervision for relation extraction in zero-shot scenarios.

Subsequently, \citet{rajpurkar-etal-2018-know} introduced a crowdsourcing process to annotate unanswerable questions, resulting in the creation of the SQuAD 2.0 dataset. This dataset later inspired similar works in other languages, such as French \cite{heinrich-etal-2022-fquad2} and Vietnamese \cite{vannguyen2022vlsp}. However, recent research has indicated that models trained on SQuAD 2.0 exhibit poor performance on out-of-domain samples \cite{sulem-etal-2021-know-dont}. 

Furthermore, apart from the adversarially-crafted unanswerable questions introduced by \citet{rajpurkar-etal-2018-know}, Natural Question \cite{kwiatkowski-etal-2019-natural} and Tydi QA \cite{clark-etal-2020-tydi} present  more naturally constructed unanswerable questions. While recent language models surpass human performances on adversarial unanswerable questions of SQuAD 2.0, natural unanswerable questions in Natural Question and Tidy QA  remain a challenging task \cite{asai-choi-2021-challenges}.

In a prior work, \citet{zhu-etal-2019-learning} introduce a pair-to-sequence model for generating unanswerable questions. However, this model requires a substantial number of high-quality unanswerable questions from SQuAD 2.0 during the training phase to generate its own high-quality unanswerable questions. Therefore, the model introduced by \citet{zhu-etal-2019-learning} cannot be applied on the HotpotQA dataset for generating high-quality unanswerable questions. In contrast, although our \textit{AGent} pipeline cannot generate questions from scratch, it distinguishes itself by its ability to create high-quality unanswerable questions without any preexisting sets of unanswerable questions. 

\subsection{Robustness of MRC Models}
The evaluation of Machine Reading Comprehension (MRC) model robustness typically involves assessing their performance against adversarial attacks and distribution shifts. The research on adversarial attacks in MRC encompasses various forms of perturbations \cite{si-etal-2021-benchmarking}. These attacks include replacing words with WordNet antonyms \cite{jia-liang-2017-adversarial}, replacing words with words having similar representations in vector space \cite{jia-liang-2017-adversarial}, substituting entity names with other names \cite{yan-etal-2022-robustness}, paraphrasing question \cite{gan-ng-2019-improving, ribeiro-etal-2018-semantically}, or injecting distractors into sentences \cite{jia-liang-2017-adversarial, Zhou_Luo_Wu_2020}. Recently, multiple innovative studies have focused on enhancing the robustness of MRC models against adversarial attacks \cite{chen-etal-2022-rationalization, zhang-etal-2023-learning, tran-etal-2023-impacts}.

On the other hand, in the research line of robustness under distribution shift, researchers study the robustness of models in out-of-domains settings using test datasets different from training dataset \cite{pmlr-v119-miller20a, fisch-etal-2019-mrqa, sen-saffari-2020-models}.

\section{Tasks and Models}

In the task of EQA, models are trained to extract a list of prospective outputs (answers), each accompanied by a probability (output of softmax function) that represents the machine's confidence in the answer's accuracy. When the dataset includes unanswerable questions, a valid response in the extracted list can be an ``empty'' response indicating that the question is unanswerable. The evaluation metric commonly used to assess the performance of the EQA system is the F1-score, which measures the average overlap between the model's predictions and the correct answers (gold answers) in the dataset. For more detailed information, please refer to the work by \citet{rajpurkar-etal-2016-squad}.
\subsection{Datasets}
In our work, we utilize three datasets: SQuAD \cite{rajpurkar-etal-2016-squad, rajpurkar-etal-2018-know}, HotpotQA \cite{yang-etal-2018-hotpotqa}, and Natural Questions \cite{kwiatkowski-etal-2019-natural}. In the SQuAD dataset, each question is associated with a short paragraph from Wikipedia. HotpotQA is a dataset designed for multi-hop reasoning question answering where each question requires reasoning over multiple supporting paragraphs. Additionally, the Natural Questions dataset comprises real queries from the Google search engine, and each question is associated with a Wikipedia page.


 \subsection{Models}
We employ three transformer-based models in our work: BERT \cite{devlin-etal-2019-bert}, RoBERTa \cite{DBLP:journals/corr/abs-1907-11692}, and SpanBERT \cite{joshi-etal-2020-spanbert}. \textbf{BERT} is considered the pioneering application of the Transformer model architecture \cite{10.5555/3295222.3295349}. BERT is trained on a combination of English Wikipedia and BookCorpus using masked language modeling and next-sentence prediction as pre-training tasks. Later, a replication study by \citet{DBLP:journals/corr/abs-1907-11692} found that BERT was significantly under-trained. \citet{DBLP:journals/corr/abs-1907-11692} built \textbf{RoBERTa} from BERT by extending the pre-training time and increasing the size of the pre-training data. \citet{joshi-etal-2020-spanbert} developed \textbf{SpanBERT} by enhancing BERT's ability to represent and predict text spans by masking random contiguous spans and replacing NSP with a span boundary objective.

Each of these three models has two versions: base and large. Our study uses all six of these models.
\section{Automatically Creating Unanswerable Questions}
\begin{figure*}[ht]
\centering
\includegraphics[width =13cm]{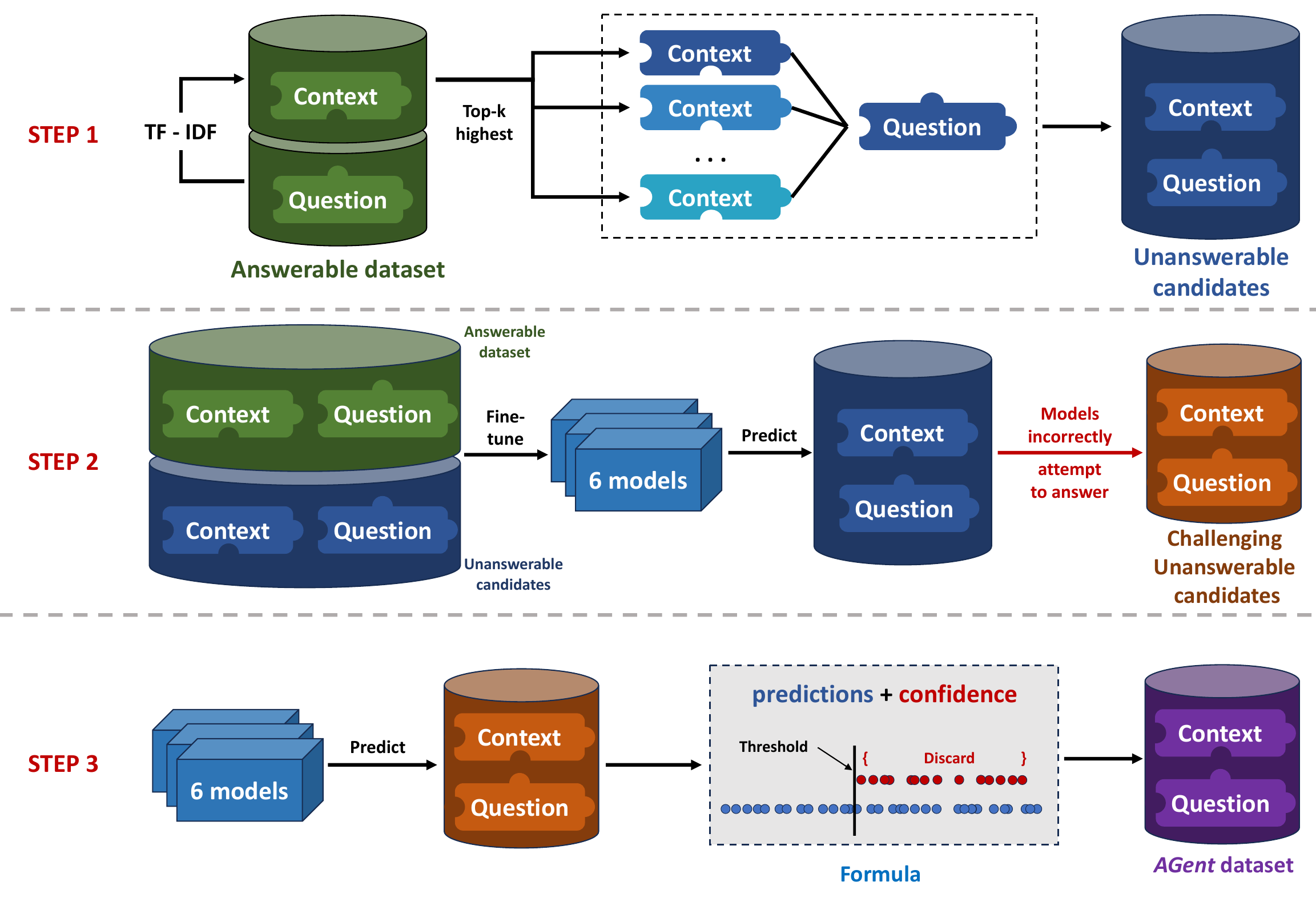}
\caption{The \textit{AGent} pipeline for generating challenging high-quality unanswerable questions in Extractive Question Answering given a dataset with answerable questions. The six models used in this pipeline are the base and large versions of BERT, RoBERTa, and SpanBERT. In step 3 of the pipeline, the \textcolor{blue}{blue dots} represent the calculated values (using formula discussed in \textsection \ref{sec:pipeline}) for unanswerable questions, while the \textcolor{red}{red dots} represent the calculated values for answerable questions. The threshold for discarding questions from the final extracted set of unanswerable questions is determined by finding the minimum value among all answerable questions. Any question with a calculated value greater than the threshold will not be included in our final extracted set.}
\label{fig:pipeline}
\end{figure*}

\subsection{Criteria}
\label{sec:desiderata}
In order to guarantee the quality of our automatically created unanswerable questions, we design our pipeline to adhere to the following criteria:

\textbf{Relevance. } The created unanswerable questions should be closely related to the subject matter discussed in the corresponding paragraph. This criterion ensures that the unanswerability of the question is not easily recognizable by simple heuristic methods and that the created question ``makes sense'' regarding the provided context.

\textbf{Plausibility. } Our pipeline also ensures that the created unanswerable questions have at least one plausible answer. For instance, when considering a question like ``What is the name of one algorithm useful for conveniently testing the primality of large numbers?'', there should exist a plausible answer in the form of the name of an algorithm in Mathematics that is closely linked to the primality within the corresponding context. See Figure \ref{fig:example} for an example showcasing an unanswerable question with strong plausible answer(s).

\textbf{Fidelity. } Our pipeline adds an additional step to ensure a minimal rate of error or noise in the set of automatically created unanswerable questions. It is important that the newly created questions are genuinely unanswerable. This quality control measure bolsters the reliability of the pipeline. The effectiveness of this step is verified in the study in Section \ref{sec:human-reviewing}.

\subsection{\textit{AGent} Pipeline}
\label{sec:pipeline}
Figure \ref{fig:pipeline} provides a summary of all the steps in the \textit{AGent} pipeline for automatically creating unanswerable questions corresponding to each dataset of answerable questions. Our proposed \textit{AGent} pipeline consists of three steps which align with the three criteria discussed in Section \ref{sec:desiderata}:

\subsubsection*{Step 1}
\textbf{Matching questions with new contexts. }In the EQA task, the input consists of a question and a corresponding context. By matching the question with a new context that differs from the original context, we can create a new question-context pair that is highly likely to be unanswerable. This step prioritizes the criterion of \textbf{relevance}. We employ the term frequency–inverse document frequency (TF-IDF) method to retrieve the $k$ most relevant paragraphs from the large corpus containing all contexts from the original dataset (while obviously discarding the context that was originally matched with this question). The outcome of this step is a set of \textbf{unanswerable candidates}. It's important to note that the unanswerable candidates created in this step may includes some answerable questions, and these answerable questions will be filtered out in step 3 of the pipeline.

\subsubsection*{Step 2}
\textbf{Identifying hard unanswerable questions. }In this step, we give priority to both the \textbf{relevance} and \textbf{plausibility} criteria. We aim to identify unanswerable questions with a highly relevant corresponding context and at least one strong plausible answer. To achieve this, we leverage the concept of adversarial filtering where the adversarial model(s) is applied to filter out easy examples \cite{yang-etal-2018-hotpotqa,zellers-etal-2018-swag,zhang2018record}. 

We first fine-tune six models using a dataset comprising answerable questions from the original dataset and randomly selected unanswerable candidates. We acknowledge that some unanswerable questions in this training set may be answerable. Nevertheless, the percentage of answerable questions among the unanswerable candidates is minimal and within an acceptable range (Appendix \ref{appendix:step2}). To ensure training integrity, we then exclude all unanswerable questions utilized for training these six models from the set of unanswerable candidates. Then, we employ the six fine-tuned models to evaluate the difficulty of each sample in the set of unanswerable candidates. If at least two of the six models predict that a given question is answerable, we consider it to be a challenging unanswerable question and include it in our set of \textbf{challenging unanswerable candidates}. 

\subsubsection*{Step 3}
\textbf{Filtering out answerable questions.} The set of challenging unanswerable questions consists of questions that at least two out of the six models predict as answerable. Consequently, there may be a considerable percentage of questions that are indeed answerable. Therefore, this specific step in our pipeline aims to ensure the \textbf{fidelity} of the \textit{AGent} pipeline, ensuring that all questions created by our pipeline are genuinely unanswerable. We leverage the predicted answers and confidence scores from the six deployed models in the previous step to achieve this. Subsequently, we devise a filtering model with four inputs: $c_a$, representing the cumulative confidence scores of the models attempting to answer (or predicting as answerable); $c_u$, representing the cumulative confidence scores of the models not providing an answer (or predicting as unanswerable); $n_a$, denoting the number of models attempting to answer; and $n_u$, indicating the number of models not providing an answer. The output of this filtering model is a value $V(q)$ for each question $q$. The filtering models must be developed independently for different datasets.

In order to determine the filtering threshold and develop the filtering model, we manually annotate $200$ challenging unanswerable candidates from each dataset. The filtering threshold is established by identifying the minimum value $V(q_a)$ where $q_a$ represents an answerable question from our annotated set. This approach ensures a precision of $100\%$ in identifying unanswerable questions on the annotated 200 questions. The filtering model then acts to minimize the number of false positives (number of unanswerable candidates that are answerable) at the expense of tossing out some candidate questions that are unanswerable. However, as the filtering model is applied on unseen challenging unanswerable candidates, the precision of the filtering model in this step would not be $100\%$ as on the $200$ muanually annotated samples. Therefore, in next section, we use human experts to evaluate the precision exhibited by the filtering model.

Further details for the \textit{AGent} pipeline are outlined in Appendix \ref{appendix:detailed-agent}.



\subsection{Human Reviewing}
\label{sec:human-reviewing}
This section presents our methodology for evaluating the data quality of unanswerable questions automatically created by \textit{AGent}.
\begin{table}[ht]
\centering
\resizebox{7cm}{!}{%
\begin{tabular}{clcc}
\hline
 & \multicolumn{1}{c}{} & \textbf{\begin{tabular}[c]{@{}c@{}}Phase\\ 1\end{tabular}} & \textbf{\begin{tabular}[c]{@{}c@{}}Phase\\ 2\end{tabular}} \\ \hline
\multirow{2}{*}{\textbf{\begin{tabular}[c]{@{}c@{}}SQuAD\\ \textit{AGent}\end{tabular}}} & Fleiss' Kappa & $0.76$ & $0.95$ \\
 & Data Error & $0.10$ & \textbf{0.06} \\ \hline
\multirow{2}{*}{\textbf{\begin{tabular}[c]{@{}c@{}}HotpotQA\\ \textit{AGent}\end{tabular}}} & Fleiss' Kappa & $0.83$ & $0.97$ \\ 
 & Data Error & $0.09$ & \textbf{0.05} \\ \hline
\end{tabular}
}
\caption{The Fleiss' Kappa score and \textit{AGent} data error for the annotations collected from human experts after two distinct phases.}
\label{tab:human-reviewing}
\end{table}

\begin{table*}[ht]
\centering
\resizebox{15cm}{!}{%
\begin{tabular}{|c|ccc|ccc|cc|}
\hline
\multicolumn{1}{|r|}{\textbf{\textit{Test}} $\rightarrow$} & \multicolumn{3}{c|}{\textbf{SQuAD}} & \multicolumn{3}{c|}{\textbf{HotpotQA}} & \multicolumn{2}{c|}{\textbf{Natural Questions}} \\
\multicolumn{1}{|l|}{\textbf{\textit{Train}} $\downarrow$} & answerable & unanswerable & \textit{AGent} & answerable & unanswerable & \textit{AGent} & answerable & \multicolumn{1}{c|}{unanswerable} \\ \hline
\begin{tabular}[c]{@{}c@{}}SQuAD \\ 2.0\end{tabular} & $84.55_{\pm 3.43}$ & \textbf{79.16} $_{\pm 5.16}$ & $49.38_{\pm 5.21}$ & $51.05_{\pm 5.15}$ & $86.28_{\pm 2.68}$ & $58.98 _{\pm 4.64}$& \textbf{44.30}$_{\pm 6.36}$ & $60.55_{\pm 12.95}$ \\ \hline
\begin{tabular}[c]{@{}c@{}}SQuAD\\ \textit{AGent}\end{tabular} & \textbf{86.96$_{\pm 1.86}$} & $29.63_{\pm 3.97}$ & $81.38_{\pm 4.52}$ & $63.26_{\pm 2.88}$ & $90.01_{\pm 2.40}$ & $50.61_{\pm 5.56}$ & $41.05_{\pm 6.81}$ & $78.66_{\pm 13.22}$ \\ \hline
\begin{tabular}[c]{@{}c@{}}HotpotQA\\ \textit{AGent}\end{tabular} & $59.06_{\pm 6.26}$ & $46.13_{\pm 3.46}$ & \textbf{87.61}$_{\pm 2.72}$ & \textbf{77.75}$_{\pm 1.92}$ & \textbf{99.70}$_{\pm0.06}$ & \textbf{95.94}$_{\pm 2.13}$ & $24.11_{\pm 7.04}$ & \textbf{84.20}$_{\pm 11.37}$ \\ \hline
\end{tabular}
}
\caption{Performance of 6 models fine-tuned on SQuAD 2.0, SQuAD \textit{AGent}, and HotpotQA \textit{AGent} datasets evaluated on SQuAD, HotpotQA, and Natural Questions. Each entry in the table is the mean and standard deviation of the F1 scores of the six MRC models. The left column indicates the dataset used to train the six MRC models. The top row indicates the dataset used to test the six MRC models. \textit{AGent} refers to the unanswerable questions generated using the \textit{AGent} pipeline. For a more detailed version of this table, refer to Table \ref{tab:detailed-result}.}
\label{tab:main-experiment}
\end{table*}

We use three experts to validate 100 random unanswerable questions from each development set of SQuAD \textit{AGent} and HotpotQA \textit{AGent}. In order to prevent an overwhelming majority of unanswerable questions in our annotation set, which could potentially undermine the integrity of the annotation, we incorporate 20 manually annotated answerable questions during step 3 of the pipeline. Consequently, we provide a total of 120 questions to each expert for each set.

The process of expert evaluation involves two distinct phases. During the first phase, each of the three experts independently assesses whether a given question is answerable and provides the reasoning behind their annotation. In the second phase, all three experts are presented with the reasons provided by the other experts for any conflicting samples. They have the opportunity to review and potentially modify their final set of annotations based on the reasons from their peers.


We observe that the annotations provided by our three experts demonstrate exceptional quality. Table \ref{tab:human-reviewing} presents the Fleiss' Kappa score \cite{Fleiss_1971} for our three experts after the completion of both phases, as well as the error rate of the \textit{AGent} development set. Notably, the Fleiss' Kappa score in phase 1 is remarkably high ($0.76$ on SQuAD \textit{AGent}, and $0.83$ on HotpotQA \textit{AGent}), suggesting that the annotations obtained through this process are reliable. Besides, after the second phase, all three experts agree that the $20$ answerable questions we include in the annotation sets are indeed answerable.

As demonstrated in Table \ref{tab:human-reviewing}, the high-quality annotations provided by three experts indicate an exceptionally low error rate for the unanswerable questions created using \textit{AGent} ($6\%$ for SQuAD and $5\%$ for HotpotQA). For comparison, this error rate is slightly lower than that of SQuAD 2.0, a dataset annotated by humans.
\section{Experiments and Analysis}
We now shift our attention from the \textit{AGent} pipeline to examining the effectiveness of our \textit{AGent} questions in training and benchmarking EQA models. 
\subsection{Training Sets}
The models in our experiments are trained using SQuAD 2.0, SQuAD \textit{AGent}, and HotpotQA \textit{AGent}. It is important to note that the two \textit{AGent} datasets includes all answerable questions from the original datasets and \textit{AGent} unanswerable questions.
\subsection{Testing Sets}
In our experiments, we use eight sets of EQA questions as summarized in Table \ref{tab:main-experiment}. In addition to two sets of \textit{AGent} unanswerable questions, we also incorporate the following six types of questions.

\textbf{SQuAD.  }We use all \textbf{answerable} questions from SQuAD 1.1. We use all \textbf{unanswerable} questions from SQuAD 2.0.

\textbf{HotpotQA.  } In preprocessing \textbf{answerable} questions in HotpotQA, we adopt the same approach outlined in MRQA 2019 \cite{fisch-etal-2019-mrqa} to convert each dataset to the standardized EQA format. Specifically, we include only two supporting paragraphs in our answerable questions and exclude distractor paragraphs.

In preprocessing \textbf{unanswerable} questions in HotpotQA, we randomly select two distractor paragraphs provided in the original HotpotQA dataset, which are then used as the context for the corresponding question.

\textbf{Natural Questions (NQ). }In preprocessing \textbf{answerable} questions in NQ, we again adopt the same approach outlined in MRQA 2019 to convert each dataset to the standardized EQA format. This format entails having a single context, limited in length. Specifically, we select examples with short answers as our answerable questions and use the corresponding long answer as the context.

For \textbf{unanswerable} questions in NQ, we select questions with no  answer and utilize the entire Wikipedia page, which is the input of original task of NQ, as the corresponding context. However, in line with the data collection process of MRQA 2019, we truncate the Wikipedia page, limiting it to the first 800 tokens.
\begin{table*}[ht]
\centering
\resizebox{15cm}{!}{%
\begin{tabular}{ll|cc}
\hline
 &  & \begin{tabular}[c]{@{}c@{}}\textbf{SQuAD}\\ \textbf{2.0}\\$\%$\end{tabular} & \begin{tabular}[c]{@{}c@{}}\textbf{SQuAD}\\ \textbf{\textit{AGent}}\\$\%$\end{tabular} \\ \hline
\begin{tabular}[c]{@{}l@{}}Insufficient \\ context for \\ question\end{tabular} & \begin{tabular}[c]{@{}l@{}}Murray survives and , in front of the RGS trustees , accuses Fawcett of abandoning him in \\ the jungle . Fawcett elects to resign from the society rather than apologize . World War I \\ breaks out in Europe , and Fawcett goes to France to fight . Manley dies in the trenches at \\ the Battle of the Somme , and Fawcett is temporarily blinded in a chlorine gas attack . Jack , \\ Fawcett 's eldest son -- who had long accused Fawcett of abandoning the family -- reconciles \\ with his father as he recovers .\\ \textbf{Question}: who dies in \textcolor{red}{the lost city of z}?\end{tabular} & $54$ & $63$ \\ \hline
\begin{tabular}[c]{@{}l@{}}typographical\\errors of key \\ words\end{tabular} & \begin{tabular}[c]{@{}l@{}}\textcolor{red}{Gimme Gimme Gimme} has broadcast three series and 19 episodes in total . The first series \\ premiered on BBC Two on 8 January 1999 and lasted for six episodes , concluding on 12 \\ February 1999 . {[}..{]}\\ \textbf{Question}: when did \textcolor{red}{gim me gim me gim me} start?\end{tabular} & $3$ & $6$ \\ \hline
\end{tabular}
}
\caption{Examples of two types of answerable questions in Natural Questions that can pose challenges for EQA models fine-tuned solely on unanswerable questions. We conduct a survey to measure the failure rates of RoBERTa models fine-tuned on both SQuAD 2.0 and SQuAD \textit{AGent} for these question types.}
\label{tab:NQ-problem}
\end{table*}
\subsection{Main Results}
Table \ref{tab:main-experiment} presents the results of our experiments. Firstly, our findings demonstrate that unanswerable questions created by \textit{AGent} pose significant challenges for models fine-tuned on SQuAD 2.0, a dataset with human-annotated unanswerable questions. The average performance of the six models fine-tuned on SQuAD 2.0 and tested on SQuAD \textit{AGent} is $49.38$; the average score for testing these models on HotpotQA \textit{AGent} data is $58.98$. Notably, unanswerable questions from HotpotQA \textit{AGent} are considerably more challenging compared to their unanswerable counterparts from HotpotQA.

Secondly, models fine-tuned on two \textit{AGent} datasets exhibit comparable performance to models fine-tuned on SQuAD 2.0. On unanswerable questions from HotpotQA and NQ, models fine-tuned on \textit{AGent} datasets significantly outperform those fine-tuned on SQuAD 2.0. On answerable questions from SQuAD and HotpotQA, models fine-tuned on SQuAD \textit{AGent} also demonstrate significant improvement over those fine-tuned on SQuAD 2.0 ($86.96 - 84.55$ on SQuAD and $63.26 - 51.05$ on HotpotQA). This finding highlights the applicability of models fine-tuned on \textit{AGent} datasets to various question types.

However, on answerable questions from NQ and unanswerable questions from SQuAD 2.0,  models fine-tuned on \textit{AGent} datasets exhibit lower performance than those fine-tuned on SQuAD 2.0. On the one hand, the lower performance on unanswerable questions from SQuAD 2.0 of models fine-tuned on AGent datasets is due to the unfair comparision as models fine-tuned on AGent datasets are tested with out-of-domain samples, and models fine-tuned with SQuAD 2.0 are tested with in-domain samples.In the next section, we provide a comprehensive explanation for the lower performance on NQ answerable questions of models fine-tuned on AGent datasets.

\subsection{Analysis on Natural Questions}
\label{sec:NQ-problem}
To delve deeper into the underperformance of models fine-tuned on \textit{AGent} dataset on answerable questions of NQ, we analyze two sets of answerable questions. The first set is $100$ answerable questions that models fine-tuned on SQuAD \textit{AGent} predict as unanswerable; the second one is 100 answerable questions that models fine-tuned on SQuAD 2.0 predict as unanswerable. For the sake of simplicity, we limit our reporting in this section to the analysis of models RoBERTa-base. Our analysis uncovers two potential issues that can arise when evaluating models with answerable questions from the NQ dataset. Table \ref{tab:NQ-problem} summarizes our findings in this section.

Firstly, a considerable difference between the original NQ dataset and the NQ used in the EQA task following a prevailing approach in the research community is the difference in the provided context. While the EQA task uses the long answer as the context \cite{fisch-etal-2019-mrqa}, NQ supplies an entire Wikipedia page as the context for a given question. This difference presents a potential problem of inadequate context for answering the question. For instance, in Table \ref{tab:NQ-problem}, we observe that the long answer associated with the question ``Who dies in the lost city of z?'' fails to mention ``the lost city of z''. Using a long answer as the context causes this question unanswerable due to the insufficient context provided. We find that most answerable questions predicted as unanswerable by models fine-tuned on SQuAD 2.0 and SQuAD \textit{AGent} belong to this specific question type ($65\%$ and $76\%$ respectively). This finding highlights the potential unreliability when comparing models using the NQ dataset in the same way as it is commonly done in multiple EQA studies. This analysis forms the basis for our decision not to employ our \textit{AGent} pipeline on the NQ dataset. 

Secondly, the questions in the NQ dataset are sourced from real users who submitted information-seeking queries to the Google search engine under natural conditions. As a result, a small portion of these questions may inevitably contain typographical errors or misspellings. In our analysis, we observe that models fine-tuned on our \textit{AGent} training set tend to predict the questions of this type as unanswerable more frequently. Nevertheless, due to the relatively small proportion of questions with typographical errors in our randomly surveyed sets, we refrain from drawing a definitive conclusion at this point. Therefore, in the subsequent section, we will delve further into this matter by adopting an adversarial attack on the EQA task. 

\section{Robustness against Syntactic Variations}
In this section, we apply the adversarial attack technique TextBugger into EQA.
\subsection{TextBugger}

Our adversarial attack in this section is inspired by the TextBugger attack \cite{Li_2019}. We use black-box TextBugger in this section, which means that the attack algorithm does not have access to the gradients of the model. TextBugger generates attack samples that closely resemble the typographical errors commonly made by real users. We perform adversarial attacks on questions from the SQuAD 1.1 dataset.

Algorithm \ref{algo:textbugger} in Appendix \ref{appendix:textbugger} provides the pseudocode outlining the process of generating attacked questions. Table \ref{tab:texbugger-example} provides examples of how TextBugger generates bugs in a given token.

\begin{table}[ht]
\centering
\resizebox{\linewidth}{!}{%
\begin{tabular}{|ccccc|}
\hline
\multicolumn{1}{|c|}{\textbf{Original}} & \multicolumn{1}{c|}{Insert} & \multicolumn{1}{c|}{Delete} & \multicolumn{1}{c|}{Swap} & \begin{tabular}[c]{@{}c@{}}Substitute\\ Character\end{tabular} \\ \hline
\multicolumn{1}{|c|}{\textbf{South}} & \multicolumn{1}{c|}{Sou th} & \multicolumn{1}{c|}{Souh} & \multicolumn{1}{c|}{Souht} & S0uth \\ \hline
\multicolumn{5}{|c|}{\begin{tabular}[c]{@{}c@{}}What \textbf{\textcolor{red}{Souh}} African law \textbf{\textcolor{red}{recongized}} two \textbf{\textcolor{red}{typ es}}\\of schools?\end{tabular}} \\ \hline
\end{tabular}
}
\caption{Examples of how TextBugger generates bugs in a given token "South" and a full question after the TextBugger attack. The attacked tokens are highlighted in \textcolor{red}{red}.}
\label{tab:texbugger-example}
\end{table}

\subsection{Robustness against TextBugger}
\begin{figure}[ht]
\centering
\includegraphics[width =\linewidth]{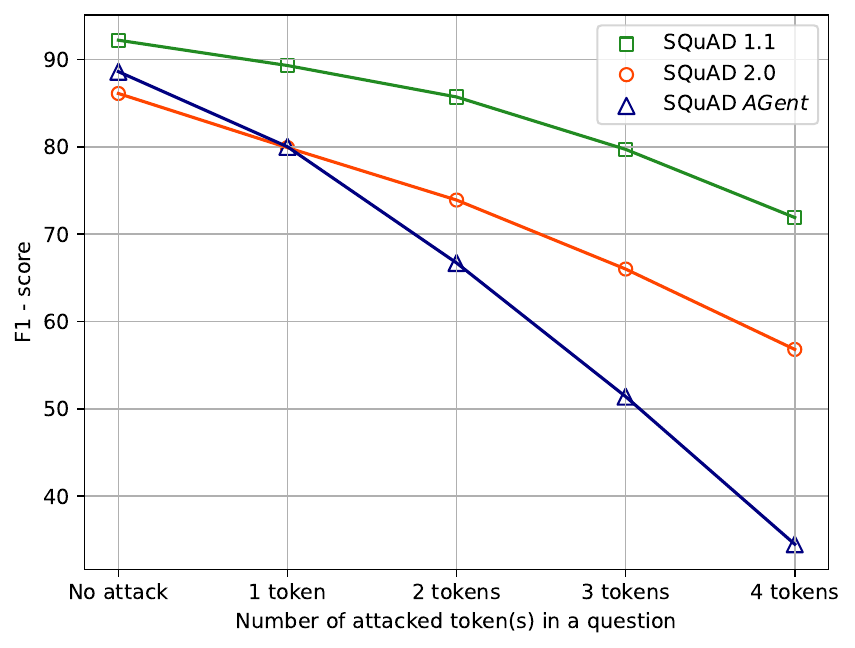}
\caption{Robustness of RoBERTa-base trained on SQuAD 1.1, SQuAD 2.0, SQuAD \textit{AGent} against TextBugger.}
\label{fig:textbugger}
\end{figure}
We investigate the impact of TextBugger attacks on models fine-tuned using different datasets, namely SQuAD 1.1, SQuAD 2.0, and SQuAD \textit{AGent}. To accomplish this, we generate attacked questions by modifying 1, 2, 3, and 4 tokens in the questions from the SQuAD 1.1 dataset.

Figure \ref{fig:textbugger} reports the performance of three models RoBERTa-base fine-tuned on SQuAD 1.1, SQuAD 2.0, and SQuAD \textit{AGent}. Firstly, we see that the performance of the model fine-tuned on SQuAD 1.1 show small decreases (from $92.2$ to $71.9$). Adversarial attack TextBugger does not present a significant challenge to the EQA model when the model is designed only to handle answerable questions. 

Secondly, we can observe that the model fine-tuned on unanswerable questions from SQuAD 2.0 demonstrates significantly better robustness compared to the model fine-tuned on SQuAD \textit{AGent} ($86.1 - 56.8$ compared to $88.6 - 34.5$). This finding confirms our initial hypothesis that the lower performance of models fine-tuned on \textit{AGent} datasets for answering questions in the NQ dataset is partly attributable to misspelled keywords in the questions from the NQ dataset.

\section{Conclusion and Future Works}

In this work, we propose \textit{AGent}, a novel pipeline designed to automatically generate two sets of unanswerable questions from a dataset of answerable questions. We systematically apply \textit{AGent} on SQuAD and HotpotQA to generate unanswerable questions. Through a two-stage process of human reviewing, we demonstrate that \textit{AGent} unanswerable questions exhibit a low error rate.

Our experimental results indicate that unanswerable questions generated using AGent pipeline present significant challenges for EQA models fine-tuned on SQuAD 2.0. We also demonstrate that models fine-tuned using \textit{AGent} unanswerable questions exhibit competitive performance compared to models fine-tuned on human-annotated unanswerable questions from SQuAD 2.0 on multiple test domains. The good performance of models finetuned on two \textit{AGent} datasets with different characteristics, SQuAD \textit{AGent} and HotpotQA \textit{AGent}, demonstrate the utility of \textit{AGent} in creating high-quality unanswerable questions and its potential for enhancing the performance of EQA models.

Furthermore, our research sheds light on two potential issues when utilizing EQA models designed to handle both answerable and unanswerable questions. Specifically, we identify the problems of insufficient context and typographical errors as considerable challenges in this context. In calling for further study on typographical errors, we propose the inclusion of the TextBugger adversarial attack in EQA. Our analysis reveals that TextBugger presents a novel challenge for EQA models designed to handle both answerable and unanswerable questions. It is important to address this challenge comprehensively before the real-world deployment of EQA models. By acknowledging and effectively tackling the influence of typographical errors, we can enhance the robustness and reliability of EQA models in practical applications.
\section*{Limitations}
We acknowledge certain limitations in our work. Firstly, our study primarily focuses on evaluating the pipeline using multiple pre-trained transformers-based models in English, which can be prohibitively expensive to create, especially for languages with limited resources. Furthermore, given the empirical nature of our study, there is no guarantee that all other transformer-based models or other deep neural networks would demonstrate the same level of effectiveness when applied in the \textit{AGent} pipeline. Consequently, the impact of the AGent pipeline on low-resource languages may be challenged due to this limitation. Potential future research could complement our findings by investigating the effectiveness of implementing \textit{AGent} pipeline in other languages.

Secondly, our analysis does not encompass a comprehensive examination of the models' robustness against various types of adversarial attacks in EQA when fine-tuned on \textit{AGent} datasets. We believe that such an analysis is crucial in determining the effectiveness of the \textit{AGent} pipeline in real-world applications, and its absence deserves further research.

Finally, our study has not discussed underlying factors for the observed phenomenon: a model fine-tuned on SQuAD AGent is less robust against TextBugger attack than its peer model fine-tuned on SQuAD 2.0. The study in this direction requires remarkably intricate investigation, which we deem beyond the scope of our present research. We leave this for our future work where we will propose our hypotheses that may shed light on this phenomenon and potential solutions to improve the robustness of EQA models against TextBugger.
\bibliography{anthology,custom}
\bibliographystyle{acl_natbib}

\newpage
\mbox{}
\newpage
\appendix
\section{\textit{AGent} on SQuAD and HotpotQA}
\label{appendix:detailed-agent}
\begin{table}[ht]
\centering
\resizebox{6cm}{!}{
\begin{tabular}{ccc}
\hline
 & \textbf{\begin{tabular}[c]{@{}c@{}}SQuAD\\ \textit{AGent}\end{tabular}} & \textbf{\begin{tabular}[c]{@{}c@{}}HotpotQA\\ \textit{AGent}\end{tabular}} \\ \hline
\begin{tabular}[c]{@{}c@{}}Unanswerable\\ Candidates\end{tabular} & $975,520$ & $1,800,550$ \\ \hline
\begin{tabular}[c]{@{}c@{}}Challenging\\ Candidates\end{tabular} & $89,432$ & $41,755$ \\ \hline
\textit{AGent} & $50,404$ & $27,840$ \\ \hline
\end{tabular}
}
\caption{Statistics of SQuAD \textit{AGent} and HotpotQA \textit{AGent} after each step of the \textit{AGent} pipeline. }
\label{tab:pipeline-stat}
\end{table}

\subsection{Generate Unanswerable Candidates}

\textbf{SQuAD. }In order to generate unanswerable candidates from questions in SQuAD 1.1, we employ bigram TF-IDF, using the question as the query, \cite{chen-etal-2017-reading} to retrieve the top-$10$ highest contexts from dataset SQuAD 1.1. Additionally, our algorithm includes a step to ensure that the set of top-$10$ highest TF-IDF scored contexts does not include the original context corresponding to the question. As a result, \textit{AGent} generates $975,520$ unanswerable candidates from SQuAD 1.1.

\textbf{HotpotQA. }In constructing benchmark settings for HotpotQA, \citet{yang-etal-2018-hotpotqa} employ bigram TF-IDF, using the question as the query, to retrieve eight paragraphs from Wikipedia as distractors. \citet{yang-etal-2018-hotpotqa} then mix these distractors with the two gold paragraphs (the ones used to collect the question and answer). We then generate unanswerable candidates from questions in HotpotQA by combining every two distractors from HotpotQA. Consequently, \textit{AGent} generates $1,800,550$ unanswerable candidates from HotpotQA.
\subsection{Identifying Challenging Unanswerable Candidates}
\label{appendix:step2}
Before using unanswerable candidates for fine-tuning the six adversarial models, we manually annotate 100 unanswerable candidates from each set of HotpotQA and SQuAD. After the manual annotation, we have $1$ answerable question from the set of SQuAD and $2$ from the set of HotpotQA. As the error rate from SQuAD 2.0 is $7\%$, we consider the error rate in unanswerable candidates is within the acceptable range for fine-tuning the six adversarial models.

In order to fine-tune adversarial models for identifying challenging unanswerable candidates, we randomly select a set of unanswerable questions from the set of unanswerable candidates from the previous step. Here, we adopt the ratio of answerable over unanswerable of SQuAD 2.0. As a result, the training set in this step for SQuAD consists of $87,599$ answerable and $43,799$ unanswerable questions; that for HotpotQA consists of $58,525$ answerable and $29,262$ unanswerable questions.

After step 2 of \textit{AGent}, we have $89,432$ and $41,755$ challenging candidates on SQuAD and HotpotQA, respectively.
\subsection{Filtering Model}
We employ a model with the following formula to classify questions as answerable or unanswerable:
$$V(q) = c_a \cdot \alpha^{n_a} - c_u \cdot \beta^{n_u}$$
In our model, we have four inputs and two adjustable parameters. Firstly, $c_a$ and $c_u$ represent the total confidence scores of the models attempting to answer (or predict as answerable) and the models not providing an answer (or predict as unanswerable), respectively. Additionally, $n_a$ and $n_u$ denote the number of models attempting to answer and the number of models not providing an answer, respectively. The parameters $\alpha$ and $\beta$ are tunable parameters. 

In order to tune the filtering model, we manually annotate $200$ questions from each set challenging unanswerable candidates. We define the difficulty level for a particular question as the number of models predicting it as answerable. Consequently, our sets of challenging unanswerable candidates encompass five difficulty levels (from $2$ to $6$). From each level, we randomly choose $40$ questions for manual annotation.

Next, we employ grid search with the step size of $0.01$ to tune for the parameters $\alpha$ and $\beta$ within the range of $(0, 2]$ with the objective of maximizing the recall of unanswerable questions, aiming to include as many unanswerable questions as possible in our final dataset. As a result, on SQuAD, we have $\alpha = 0.64$ and $\beta = 0.69$; on HotpotQA, we have $\alpha = 0.52$ and $\beta = 0.94$. After going through the filtering model, SQuAD AGent has $50,404$ unanswerable questions; HotpotQA AGent has $27,840$.

\section{Details for Models Training}
The input of a question-context pair into the pre-trained model is in the form of \textit{<Question>[SEP]<Context>}, with \textit{[SEP]} as a special token of pre-trained tokenizer accompanying the pre-trained model. After getting embeddings for each token, we feed its final embedding into a start and end token classifier. After taking the dot product between the output embeddings and the classifier's weights, we apply the softmax activation to produce a probability distribution over all words. The word with the highest probability after the start classifier will be predicted as the start of the answer span.
\begin{table}[ht]
\centering
\resizebox{7cm}{!}{%
\begin{tabular}{lcc}
\hline
 & total samples & \# unanswerable \\ \hline
\begin{tabular}[c]{@{}l@{}}\textbf{SQuAD} \\ \textbf{\textit{Adversarial}}\end{tabular} & 130,319 & 43,439 \\ \hdashline
\begin{tabular}[c]{@{}l@{}}\textbf{HotpotQA} \\ \textbf{\textit{Adversarial}}\end{tabular} & 87,787 & 29,262 \\ \hline
\begin{tabular}[c]{@{}l@{}}\textbf{SQuAD} \\ \textbf{\textit{AGent}}\end{tabular} & 135, 615 & 48, 016 \\ \hdashline
\begin{tabular}[c]{@{}l@{}}\textbf{HotpotQA}\\ \textbf{\textit{AGent}}\end{tabular} & 83, 589 & 25, 064 \\ \hdashline
\textbf{SQuAD 2.0} & 130,319 & 43,498 \\ \hline
\end{tabular}
}
\caption{Data statistics of all training sets used in this paper. Adversarial datasets refer to training sets for the adversarial models in Step 2.}
\label{tab:training-stat}
\end{table}
\begin{table}[ht]
\centering
\resizebox{\linewidth}{!}{%
\begin{tabular}{lccc}
\hline
 & \textbf{SQuAD} & \textbf{HotpotQA} & \textbf{NQ} \\ \hline
Answerable & 11,873 & 5,901 & 12,836 \\
Unanswerable & 5,945 & 5,918 & 2,331 \\
\textit{\textbf{AGent}} & 2,217 & 2,776 & $-$ \\ \hline
\end{tabular}
}
\caption{Data statistics of all testing sets used in this paper. \textit{AGent} refers to the unanswerable questions generated using the \textit{AGent} pipeline.}
\label{tab:testing-stat}
\end{table}

Table \ref{tab:training-stat} provides the statistics for all training sets in this paper. Table \ref{tab:testing-stat} provides the statistics for all testing sets in this paper.

We train all models with batch size of 8 for 2 epochs. The maximum sequence length is set to 384 tokens. We use the AdamW optimizer \cite{loshchilov2018decoupled} with an initial learning rate of $2 \cdot 10^{-5}$, and $\beta_1 = 0.9$, $\beta_2 = 0.999$. We use a single NVIDIA GeForce RTX 3080 for training and evaluating models.

\section{Detailed Results of Main Experiments}
Table \ref{tab:detailed-result} presents a detailed version of our experiments with training six models on SQuAD 2.0, SQuAD \textit{AGent}, and HotpotQA \textit{AGent} and evaluating on SQuAD, HotpotQA, and Natural Questions.
\begin{table*}[ht]
\centering
\resizebox{\textwidth}{!}{%
\begin{tabular}{|lll|ccc|ccc|cc|}
    \hline
     &  &  & \multicolumn{3}{c|}{\textbf{SQuAD}} & \multicolumn{3}{c|}{\textbf{HotpotQA}} & \multicolumn{2}{c|}{\textbf{Natural Questions}} \\
     &  &  & answerable & unanswerable & \textit{AGent} & answerable & unanswerable & \textit{AGent} & answerable & unanswerable \\ \hline
     
    \multirow{6}{*}{\begin{tabular}[c]{@{}c@{}}\textbf{SQuAD}\\ \textbf{2.0}\end{tabular}} & \multirow{2}{*}{BERT} & base & 78.2 & 70.9 & 43.6 & 42.7 & 84.2 & 58.2 & 34.7 & 53.2 \\
     &  & large & 84.5 & 77.2 & 46.5 & 50.1 & 85.8 & 61.5 & 38.7 & 53.4 \\ \cline{2-11} 
     
     & \multirow{2}{*}{RoBERTa} & base & 84.5 & 82.5 & 54.1 & 50.0 & 88.5 & 59.6 & 45.1 & 78.7 \\
     &  & large & 85.7 & 84.6 & 57.1 & 50.4 & 89.5 & 64.9 & 46.7 & 64.7 \\ \cline{2-11} 
     
     & \multirow{2}{*}{SpanBERT} & base & 85.9 & 76.8 & 45.9 & 56.7 & 82.4 & 50.9 & 50.9 & 70.0 \\
     &  & large & 88.5 & 83.0 & 49.1 & 56.4 & 87.3 & 58.8 & 49.7 & 43.3 \\ \hline
     
    \multirow{6}{*}{\begin{tabular}[c]{@{}c@{}}\textbf{SQuAD}\\ \textbf{\textit{AGent}}\end{tabular}} & \multirow{2}{*}{BERT} & base & 83.6 & 23.6 & 77.0 & 58.1 & 86.6 & 42.0 & 30.0 & 81.2 \\
     &  & large & 86.8 & 28.2 & 82.0 & 62.8 & 91.0 & 51.6 & 36.3 & 68.2 \\ \cline{2-11} 
     
     & \multirow{2}{*}{RoBERTa} & base & 87.6 & 29.2 & 86.2 & 63.8 & 91.6 & 53.8 & 41.9 & 90.7 \\
     &  & large & 87.3 & 34.6 & 86.5 & 64.9 & 92.4 & 56.5 & 47.8 & 57.3 \\ \cline{2-11} 
     
     & \multirow{2}{*}{SpanBERT} & base & 87.2 & 28.7 & 75.6 & 63.3 & 87.4 & 45.8 & 43.2 & 89.3 \\
     &  & large & 89.3 & 33.5 & 81.0 & 66.7 & 91.1 & 54.0 & 47.1 & 85.3 \\ \hline
     
    \multirow{6}{*}{\begin{tabular}[c]{@{}c@{}}\textbf{HotpotQA}\\ \textbf{\textit{AGent}}\end{tabular}} & \multirow{2}{*}{BERT} & base & 48.2 & 45.1 & 86.3 & 74.4 & 99.6 & 92.2 & 14.2 & 98.1 \\
     &  & large & 56.6 & 45.2 & 87.9 & 77.1 & 99.7 & 96.0 & 20.0 & 98.6 \\ \cline{2-11} 
     
     & \multirow{2}{*}{RoBERTa} & base & 62.8 & 40.6 & 82.9 & 77.7 & 99.7 & 97.2 & 24.8 & 99.5 \\
     &  & large & 62.4 & 49.2 & 89.9 & 79.0 & 99.7 & 98.3 & 35.0 & 71.0 \\ \cline{2-11} 
     
     & \multirow{2}{*}{SpanBERT} & base & 58.5 & 50.4 & 90.3 & 78.3 & 99.7 & 95.0 & 23.0 & 99.2 \\
     &  & large & 65.9 & 46.3 & 88.4 & 80.0 & 99.8 & 96.8 & 27.7 & 98.8 \\ \hline
    \end{tabular}
    }

\caption{Performance of 6 models fine-tuned on SQuAD 2.0, SQuAD \textit{AGent} and HotpotQA \textit{AGent} evaluated on SQuAD, HotpotQA, and NQ. The term \textit{AGent} refers to the unanswerable questions that are generated using the \textit{AGent} pipeline.}
\label{tab:detailed-result}
\end{table*}
\section{Unanswerable Examples}
\label{appendix:unan}
Table \ref{tab:agu-example1} and \ref{tab:agu-example2} present some notable examples of unanswerable questions generated using \textit{AGent}.

\begin{table*}
\renewcommand{\arraystretch}{1.5}
    \begin{tabularx}{\textwidth}{|p{10cm}|Y|}
        \hline
        \multicolumn{1}{|c|}{ \textbf{Unanswerable questions} } & \multicolumn{1}{c|}{ \textbf{Reasons} }\\
        \hline
        \raggedright \textbf{Question}: \\
        What is the most critical resource measured to in assessing the determination of a Turing machine's ability to solve any given set of problems? \\ \textbf{Context}: \\
        Many types of Turing machines are used to define complexity classes, such as deterministic Turing machines, probabilistic Turing machines, non-deterministic Turing machines, quantum Turing machines, symmetric Turing machines and alternating Turing machines. They are all equally powerful in principle, but when resources (such as \textcolor{red}{\textbf{time or space}}) are bounded, some of these may be more powerful than others. 
        & The context provide examples for critical resources but does not specify whether these resources are most critical or not. \\
        \hline
        
        \raggedright \textbf{Question}:\\ 
        What are the specific divisors of all even numbers larger than 2?\\
        \textbf{Context}: 
        Many questions regarding prime numbers remain open, such as Goldbach's conjecture (that every even integer greater than 2 can be expressed as the sum of \textcolor{red}{\textbf{two primes}}), and the twin prime conjecture (that there are infinitely many pairs of primes whose difference is 2). {[}...{]} 
        & The context provides insights into even numbers and primes, but it does not directly specify the divisors of all even numbers larger than 2.\\

        \hline
        \raggedright \textbf{Question}:\\ 
        What is the atomic number for oxygen? \\ 
        \textbf{Context}:\\ 
        {[}...{]} Dalton assumed that water's formula was HO, giving the atomic mass of oxygen as \textcolor{red}{\textbf{8}} times that of hydrogen, instead of the modern value of about \textcolor{red}{\textbf{16}}. {[}...{]},
        & The context only mentions the atomic mass ratio between oxygen and hydrogen. It does not provide information about the atomic number of oxygen.\\


        \hline
        \raggedright \textbf{Question}:\\ 
        When did Tesla make these claims?\\ 
        \textbf{Context}:\\ 
        {[}...{]} In \textcolor{red}{\textbf{February 1912}}, an article ``Nikola Tesla, Dreamer'' by Allan L. Benson was published in World Today, in which an artist's illustration appears showing the entire earth cracking in half with the caption, "Tesla claims that in a few weeks he could set the earth's crust into such a state of vibration that it would rise and fall hundreds of feet and practically destroy civilization. A continuation of this process would, he says, eventually split the earth in two. 
        & The context only refers to an article published in February 1912 by Allan L. Benson, which discusses Tesla's claims about setting the earth's crust into vibration. However, it does not explicitly mention when Tesla made the claims.   \\
        
        \hline
    \end{tabularx}

\caption{Examples unanswerable questions in SQuAD \textit{AGent}. The spans in \textcolor{red}{red} are strong plausible answers for the corresponding questions.}
\label{tab:agu-example1}
\end{table*}


\begin{table*}
\renewcommand{\arraystretch}{1.5}
    \begin{tabularx}{\textwidth}{|p{10cm}|Y|}
        \hline
        \multicolumn{1}{|c|}{ \textbf{Unanswerable questions} } & \multicolumn{1}{c|}{ \textbf{Reasons} }\\
        \hline
        \raggedright \textbf{Question}: \\
        Keene is an unincorporated community in Wabaunsee County, Kansas, in what federal republic composed of 50 states? \\ 
        \textbf{Context}: \\
        The \textcolor{red}{\textbf{United Mexican States}} (Spanish: ``Estados Unidos Mexicanos'' ) is a federal republic composed of 31 states and the capital, Mexico City, an autonomous entity on par with the states. Newbury is an unincorporated community in Wabaunsee County, Kansas, in the United States.
        & The context mentions the United Mexican States, which is a federal republic composed of 31 states and Mexico City. However, it does not provide any information about a federal republic composed of 50 states. \\
        \hline
        
        \raggedright \textbf{Question}:\\ 
        What was the last date the creator of the NOI was seen by Elijah Muhammad?\\
        \textbf{Context}: 
        Tynnetta Muhammad [...] wrote articles and columns for the Nation of Islam (NOI) newspaper ``Muhammad Speaks''. Having worked as a secretary to Elijah Muhammad, she made it known after his death in 1975 that she was one of his widows. Elijah Muhammad [...] led the Nation of Islam (NOI) from \textcolor{red}{\textbf{1934 until his death in 1975.}} [...].
        &  The context mentions that Elijah Muhammad led the Nation of Islam from 1934 until his death in 1975, but it does not specify the exact date of the last encounter between the creator of the NOI and Elijah Muhammad.\\

        \hline
        \raggedright \textbf{Question}:\\ 
        Polk County Florida's second most populated city is home to which mall? \\ 
        \textbf{Context}:\\ 
        \textcolor{red}{\textbf{Lakeland Square Mall}} is a shopping mall located on the northern side of Lakeland, Florida in the United States. [...] It is owned and managed by Rouse Properties, one of the largest mall owners in the United States. [...]
        & The context specifically mentions Lakeland Square Mall, which is located in Lakeland, Florida, but it does not state that Lakeland is the second most populated city in Polk County.\\

        \hline
        \raggedright \textbf{Question}:\\ 
        What podcast was the cheif executive officer of Nerdist Industries a guest on? \\ 
        \textbf{Context}:\\ 
        Nerdist News [...] was founded and operated by Nerdist Industries' CEO, Peter Levin, and its CCO, Chris Hardwick. [...] Nerdist Industries was founded as a sole podcast \textcolor{red}{\textbf{(The Nerdist Podcast)}} created by Chris Hardwick but later spread to include a network of podcasts. [...] 
        & The context mentions the Nerdist Industries CEO, Peter Levin. However, the context does not provide information about a specific podcast where the CEO of Nerdist Industries was a guest.\\

        \hline
        \raggedright \textbf{Question}:\\ 
        What book provided the foundation for Masters and Johnson's research team?\\ 
        \textbf{Context}:\\ 
        \textcolor{red}{\textbf{Sheep}} is a horror novel by British author Simon Maginn, originally published in 1994 and reissued in 1997. [...] William Howell Masters (December 27, 1915 - February 16, 2001) was an American gynecologist, best known as the senior member of the Masters and Johnson sexuality research team. [...]
        & The context mentions William Howell Masters, who was a prominent member of the Masters and Johnson sexuality research team. However, it does not specify the book that served as the foundation for their research. \\
        
        \hline
    \end{tabularx}

\caption{Examples unanswerable questions in Hotpot \textit{AGent}. The spans in \textcolor{red}{red} are strong plausible answers for the corresponding questions.}
\label{tab:agu-example2}
\end{table*}
\section{TextBugger Pseudocode}
\label{appendix:textbugger}
Algorithm \ref{algo:textbugger} presents the pseudocode of the specific version of TextBugger employed in our analysis.

\begin{algorithm*}[ht]
    \caption{TextBugger EQA Attack}\label{algo:textbugger}
    \SetAlgoHangIndent{1em}
    \SetKwFunction{FMain}{TextBugger}
    \SetKwFunction{FSub}{GenerateBug}
    \SetKwFunction{FBug}{Bug}
    \SetKwProg{Fn}{Function}{:}{}
    \SetKw{assert}{Assert}
    \SetAlgoLined
    \Fn{\FMain{question, numAttack}}{
        $\mathit{attackPositions} \gets$ randomly select indices of tokens in question; \\
        \ForAll{$pos\in\mathit{attackPositions}$}
        {
            $question[pos] \gets \FSub{question[pos]}$;
        }
    }
    \Fn{\FSub{token}}{
        $newToken \gets token$ \\
        \While {$newToken \neq token$}
        {
            $bugType \gets$ randomly select Bug type;\\
            $newToken \gets \FBug{newToken, bugType}$; \\
        }
        \KwRet $newToken$
    }
\end{algorithm*}

\end{document}